\documentclass{article}

\usepackage{microtype}
\usepackage{graphicx}
\usepackage{subfigure}
\usepackage{booktabs} 

\usepackage{hyperref}

\usepackage[accepted]{icml2020}
\usepackage{tabularx}
\newcolumntype{Y}{>{\centering\arraybackslash}X}
\usepackage{longtable}
\usepackage{graphicx}
\usepackage{wrapfig}
\usepackage{hyperref}
\usepackage{url}
\usepackage{multirow}
\usepackage[colorinlistoftodos]{todonotes}
\usepackage{amsmath}
\usepackage{hhline}
\usepackage{amssymb}
\usepackage{listings}
\newcommand{\xx}{\mathbf{x}}

\newcommand{\x}{\mathbf{x}}

\newcommand{\E}{\mathbf{E}}
\newcommand{\R}{\mathcal{R}}
\newcommand{\F}{\mathcal{F}}
\newcommand{\St}{\mathbf{S}}
\newcommand{\LT}{\mathcal{L}_2}
\newcommand{\LSD}{\rm{\bf{LSD}}}

\icmltitlerunning{Learning the Stein Discrepancy}

\begin{document}

\twocolumn[
\icmltitle{Learning the Stein Discrepancy \\
for Training and Evaluating Energy-Based Models without Sampling}




\begin{icmlauthorlist}
\icmlauthor{Will Grathwohl}{to}
\icmlauthor{Kuan-Chieh Wang}{to}
\icmlauthor{J\"orn-Henrik Jacobsen}{to}
\icmlauthor{David Duvenaud}{to}
\icmlauthor{Richard Zemel}{to}
\end{icmlauthorlist}

\icmlaffiliation{to}{University of Toronto and Vector Institute, Toronto, Canada}

\icmlcorrespondingauthor{Will Grathwohl}{wgrathwohl@cs.toronto.edu}

\icmlkeywords{Machine Learning, ICML, EBM, Generative Model, Stein Discrepancy}

\vskip 0.3in
]



\printAffiliationsAndNotice{}  

\begin{abstract}
We present a new method for evaluating and training unnormalized density models.
Our approach only requires access to the gradient of the unnormalized model's log-density.
We estimate the Stein discrepancy between the data density $p(x)$
and the model density $q(x)$
defined by a vector function of the data. We parameterize this function
with a neural network and fit its parameters to maximize the discrepancy.
This yields a novel goodness-of-fit test which outperforms existing methods on high dimensional data.
Furthermore, optimizing $q(x)$ to minimize this discrepancy produces a novel method for training unnormalized models which scales more gracefully than existing methods.
The ability to both learn and compare models is a unique feature of the proposed method.
\end{abstract}

\section{Introduction}

\begin{figure}[t]   
\centering
\includegraphics[width=\linewidth, clip, trim=1.5mm 1mm 1.6mm 0.6mm]{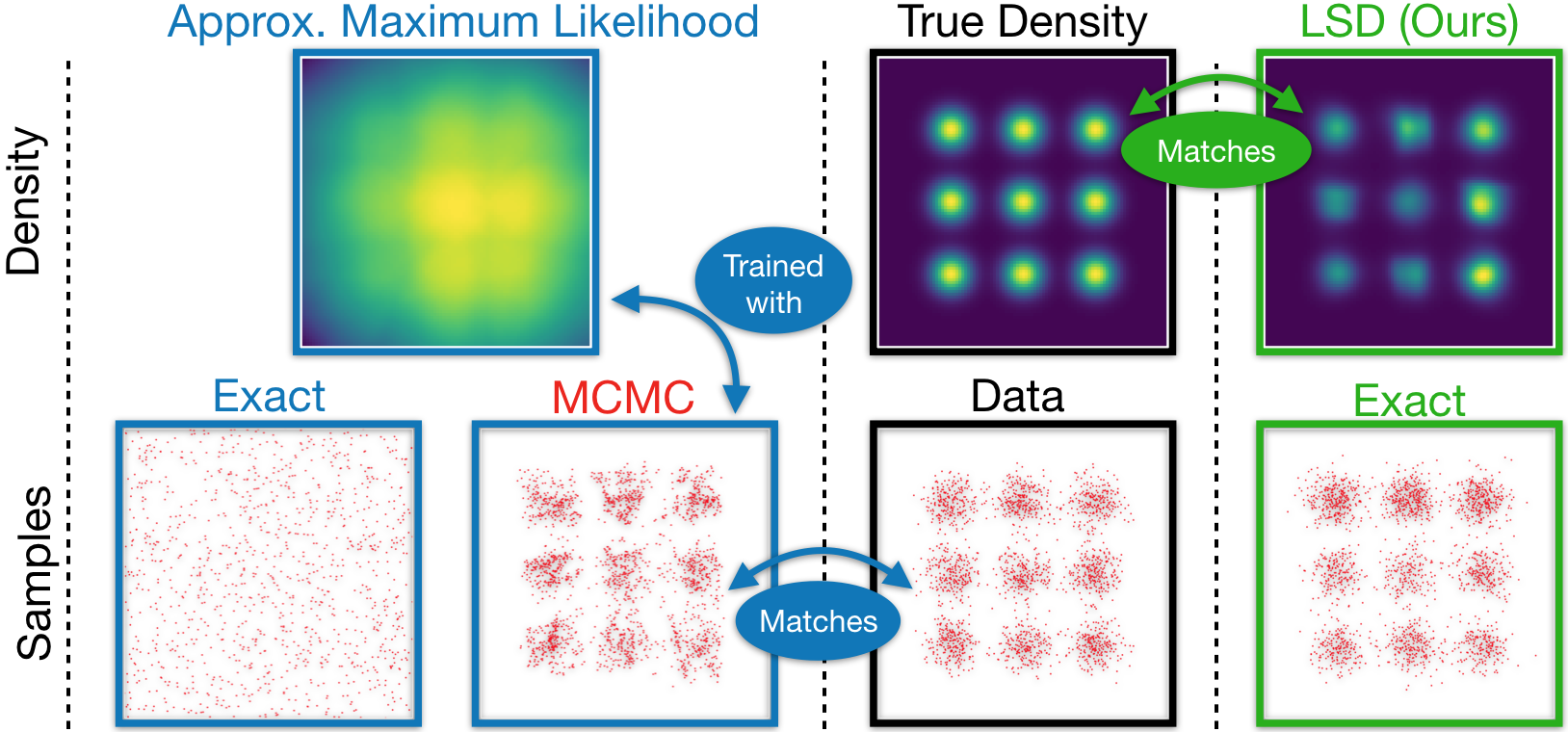}
    \caption{
    Density models trained with approximate MCMC samplers can fail to match the data density while still generating high-quality samples. Samples from approximate MCMC samplers follow a \emph{different} distribution than the density they are applied to. It is this induced distribution which is trained to match the data. In contrast, our approach $\LSD$ directly matches the model density to the data density without reliance on a sampler. 
    \vspace{-.6cm}} 
    \label{fig:3-dist}
\end{figure}

Energy-Based Models (EBMs), also known as unnormalized density models, are perhaps the most flexible way to parameterize a density.
They hinge on the observation that any density $p(x)$ can be expressed as
\begin{align}
    p(\x) = \frac{\exp(-E(\x))}{Z}~,
    \label{eq:ebm}
\end{align}
where $E: \mathbb{R}^D \rightarrow \mathbb{R}$, known as the \emph{energy function}, maps each point to a scalar, and $Z = \int_\x \exp(-E(\x))$ is the normalizing constant.

A major benefit of EBMs is that they allow maximal freedom in designing the energy function $E$. This makes it straightforward to incorporate prior knowledge about the problem, such as symmetries or domain-specific design choices, into the structure of the model.
This has made EBMs an appealing candidate for applications in physics~\citep{noe2019boltzmann}, biology~\citep{ingraham2019learning}, neuroscience~\citep{scellier2017equilibrium}, and computer vision~\citep{lecun2007energy, osadchy2007synergistic, xie2016theory, xie2019learning, xie2018learning}, to name a few.

Despite their many benefits, EBMs present a central challenge which complicates their use:
because we cannot efficiently compute the normalizing constant, we cannot compute likelihoods under our model, making training and evaluation difficult.
Much prior work on EBMs has relied on MCMC sampling techniques to estimate the likelihood (for evaluation) and its gradient (for training).
Other approaches train EBMs by finding easier-to-compute surrogate objectives which have similar optima to the maximum likelihood objective. These include Score Matching~\cite{hyvarinen2005estimation} and Noise-Contrastive Estimation~\cite{gutmann2010noise}. 

These original sampling- and score-based approaches were not able to scale 
to large, high-dimensional datasets as well as subsequently developed alternative models, such as Variational Autoencoders (VAEs)~\cite{kingma2013auto} and Normalizing Flows (NFs)~\citep{rezende2015variational}.
These approaches offer more easily scalable training, evaluation, and sampling, but do so at the cost of a more restrictive model parameterization which can lead to well-known problems like posterior collapse in VAEs~\citep{lucas2019understanding}, and the inability of NFs to model distributions with certain topological structures~\citep{falorsi2018explorations}.

Recently, a number of improvements have been made to EBM training techniques which have enabled EBMs to be trained on high-dimensional data. 
These include improvements to MCMC-based training~\citep{nijkamp2019anatomy}, Score Matching \citep{song2019generative}, and noise-contrastive approaches~\citep{gao2019flow}.
These improvements enabled new applications in domains such as protein structure prediction~\citep{ingraham2019learning, du2020energybased}, and provided benefit to fundamental problems in machine learning such as adversarial robustness, calibration, out-of-distribution detection~\citep{grathwohl2019your,du2019implicit}, and semi-supervised learning~\citep{song2018learning}.

Despite this progress, we have little insight into the quality of the models we are learning.
One solution is to indirectly measure quality of the model by evaluating performance on downstream discriminative tasks as advocated for in \citet{theis2015note}, and recently applied to EBMs by \citet{grathwohl2019your}.
Another solution is to use metrics that rely on samples generated by expensive MCMC algorithms~\citep{nijkamp2019anatomy, song2019generative, du2019implicit}. 

In this work, we develop a unified approach to training and evaluating EBMs which addresses many of the aforementioned issues.
We produce a measure of model fit that requires only an unnormalized model and data from the target distribution.
This measure is given by a neural network which is trained to estimate the Stein Discrepancy between distributions. To our knowledge, our work is the first to empirically demonstrate that a neural network can be trained to reliably estimate the Stein Discrepancy in high dimensions. The resulting evaluation and training procedures outperform previous approaches at both tasks. 



\section{Problems with Sample-Based Training and Evaluation of Energy-based Models}

EBMs and MCMC sampling have been closely intertwined since their inception.
Much prior work trains EBMs by approximating the gradient of the model likelihood with:
\begin{align}
    \frac{\partial \log p_\theta(\x)}{\partial \theta}  = \E_{\x' \sim p_\theta(\x)}\left[ \frac{\partial E_\theta(\x')}{\partial \theta} \right] - \frac{\partial E_\theta(\x)}{\partial \theta}~,
\label{eq:grad_est}
\end{align}
where $\theta$ parameterizes the model, and the expectation on the right-hand-side of Equation~\ref{eq:grad_est} is estimated using MCMC. We refer to such approaches as Approximate Maximum Likelihood (AML). When the Markov chain is seeded from training data, this approach is referred to as Contrastive Divergence (CD). 
This gradient estimator has been used in the past to train product-of-experts~\citep{hinton2002training} and Restricted Boltzmann Machines~\citep{hinton2006fast}. Recently, advances in generic gradient-based samplers~\citep{welling2011bayesian} have allowed this gradient estimator to be used to train much more large-scale models~\citep{du2019implicit, grathwohl2019your, ingraham2019learning}.

Unfortunately, unless extreme care is taken~\citep{jacob2017unbiased}, estimating an expectation with a finite Markov chain will produce biased estimates.
This leads to optimizing an objective other than the one we wish to. An MCMC sampler will only draw true samples from an unnormalized density if it is run for an infinite number of steps. When a finite number of steps are used, the distribution of the resulting samples can be arbitrarily far away from the true distribution parameterized by the model. This difference is a function of the model itself as well as the parameters of the sampler such as its step size, number of steps, and initialization distribution. This means the bias of this training objective cannot be easily quantified and is greatly complicated by the choice of sampler. This phenomenon has been explored in \citet{nijkamp2019learning} where the authors argue that training in this way actually learns an implicit sampler and not a density model. This is illustrated in Figure \ref{fig:3-dist}.



\begin{figure}[h]
		\centering
		\includegraphics[width=80mm]{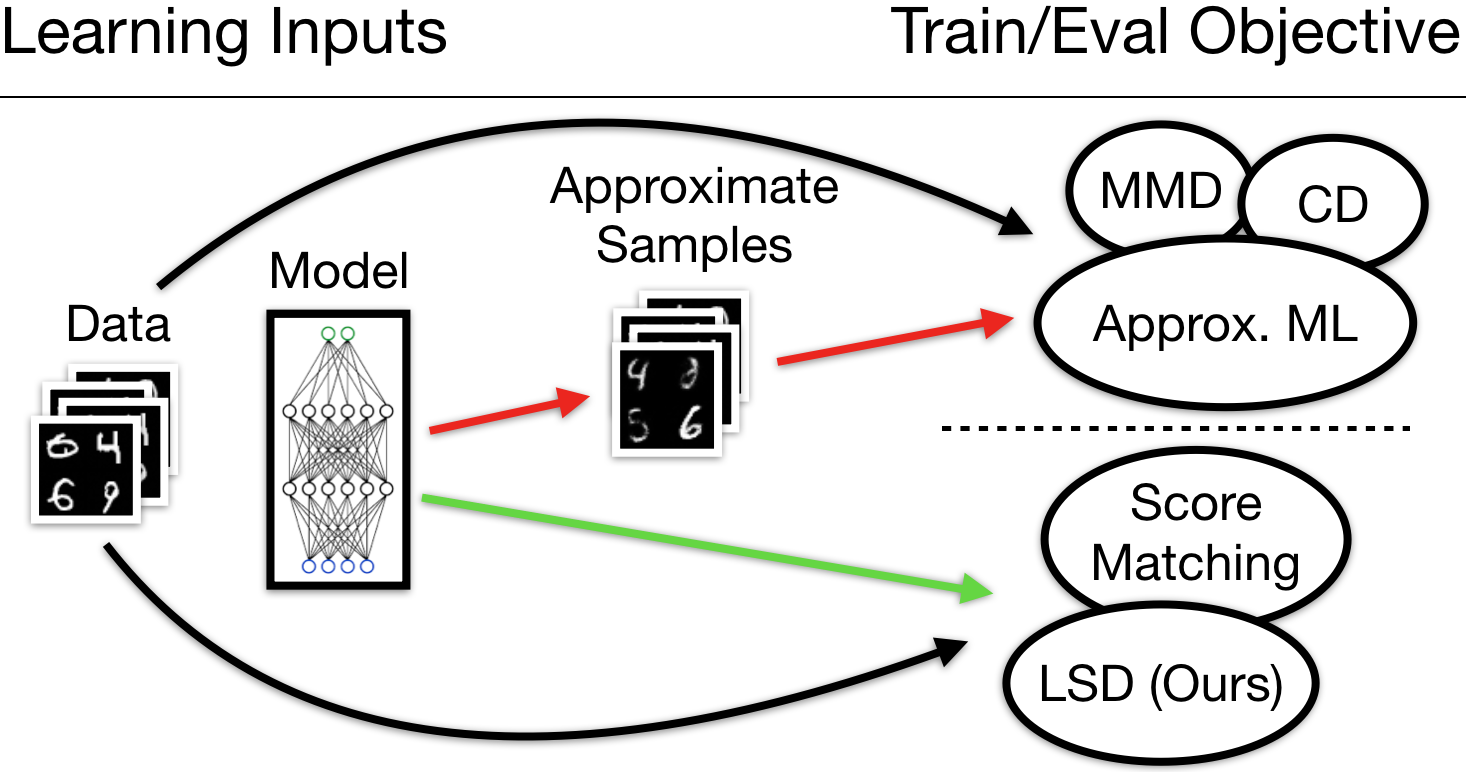} 
		\caption{Cutting out the ``middle-man'' of approximate sampling can lead to simpler training and evaluation that is tied directly to the quality of our model and is not obfuscated by the parameters of an MCMC sampler. }
\label{fig:various-training}
\end{figure}

\section{Assessing Fit Without Samples or Normalizing Constants}
To avoid the problems of sampler-based training and evaluation, we seek a measure of model fit that can be evaluated given a finite set of datapoints and an unnormalized density model.
Surprisingly, such a measure exists and is based on the following, known as Stein's Identity~\citep{stein1972bound}:
\begin{align}
    \E_{p(x)}\left[\nabla_x \log p(x)^T  f(x) + \text{Tr}\left( \nabla_x f(x)\right) \right] = 0
    \label{eq:stein_id}
\end{align}
where $f: \R^D \rightarrow \R^D$ is any function such that $\lim_{||x||\rightarrow \infty} p(x)f(x) = 0$. We refer to $f$ as the critic. 
Note that $\nabla_x \log p(x)$ can be evaluated without computing the normalizing constant of $p(x)$. If we replace the $p(x)$ inside of the expectation with a different distribution $q(x)$, then this expectation is zero 
for all $f$ 
if and only if $p = q$.

Taking $f$ to be the supremum over a class of functions $\F$, we can create a quantitative measure:
\begin{align}
    \St(p, q) = \sup_{f \in \F} \E_{p(x)}\left[\nabla_x \log q(x)^T  f(x) + \text{Tr}\left( \nabla_x f(x)\right) \right]~,
    \label{eq:stein_disc}
\end{align}
which is known as the Stein Discrepancy (SD)~\citep{gorham2017measuring}.


If $\F$ is taken to be a ball in a Reproducing Kernel Hilbert Space (RKHS), then a kernelized version of this discrepancy can be derived:
\begin{align}
    \mathbf{KSD}(p, q) = \E_{x, x' \sim p(x)}[&\nabla_x \log q(x)^T  k(x, x')   \nabla_{x'} \log q(x') \nonumber\\
    &+ \nabla_{x} \log q(x)^T \nabla_{x'} k(x, x') \nonumber\\ 
    &+ \nabla_{x} k(x, x')^T \nabla_{x'} \log q(x')\nonumber\\ 
    &+ \text{Tr}(\nabla_{x, x'} k(x, x'))]~,
    \label{eq:ksd}
\end{align}

known as the Kernelized Stein Discrepancy (KSD), which has been used to build hypothesis tests to assess goodness-of-fit for unnormalized densities~\citep{liu2016kernelized, gorham2017measuring, jitkrittum2017linear} and to learn implicit samplers for unnormalized densities~\citep{hu2018stein}.

\section{Learning the Stein Discrepancy}

The KSD allows us to circumvent the functional optimization in Equation \ref{eq:stein_disc} by using a kernel function $k(\cdot, \cdot)$. Unfortunately, it is known that methods based on distances and kernels quickly degrade in performance as dimension increases~\citep{ramdas2015decreasing}. Further, the power of tests based on these kernel methods is closely tied to their asymptotic run-time; the quadratic time test of \citet{liu2016kernelized} significantly outperforms their linear-time variant, which prevents using the best-performing approach on large datasets.

Using the Stein Discrepancy directly has the potential to address both these issues. By employing a larger and more expressive class of functions $\F$, we can produce a more discriminative measure in high dimensions. In the functional form, we can estimate the discrepancy in linear time with respect to the number of examples.

We propose to parameterize the critic with a neural network $f_\phi$ and optimize its parameters to maximize
\begin{equation}
    \mathbf{LSD}(f_\phi, p, q) = \E_{p(x)}[\nabla_x \log q(x)^T f_\phi(x) + \text{Tr}(\nabla_x f_\phi(x))]~,
    \label{eq:critic_obj}
\end{equation}
which we call the Learned Stein Discrepancy ($\LSD$). We will then use this learned discrepancy to evaluate and train unnormalized models.

\paragraph{Choosing $\F$} To estimate a Stein Discrepancy as in Equation \ref{eq:stein_disc}, we must optimize the critic over a bounded space of functions $\F$. In general, unconstrained neural networks do not fall into this category, so additional care must be taken. This can be accomplished in many ways such as with weight-clipping, or spectral normalization~\citep{miyato2018spectral}. In this work, we optimize critic networks within
\begin{align}
    \F = \{f : \E_{p(x)}[f(x)^T f(x)] < \infty\},
\end{align}
the space of functions whose squared norm has finite expectation under the data distribution. This constraint will be enforced by placing an $\mathcal{L}_2$ regularizer on our critic's output
\begin{align}
    \mathcal{R}_\lambda(f_\phi) = \lambda \E_{p(x)}[f_\phi(x)^T f_\phi(x)]
    \label{eq:reg}
\end{align}
with strength $\lambda$. Thus, our critic will be trained to maximize $\rm{\bf{LSD}}(f_\phi, p, q) - \mathcal{R}_\lambda(f_\phi)$. 

Under this class of functions, \citet{hu2018stein} prove the optimal critic takes the following form:
\begin{align}
    f_\phi(x) = \frac{1}{2\lambda}(\nabla_x \log q(x) - \nabla_x \log p(x)).
\label{eq:bound}
\end{align}

\begin{figure}[ht] 
\centering
\includegraphics[width=.2\textwidth]{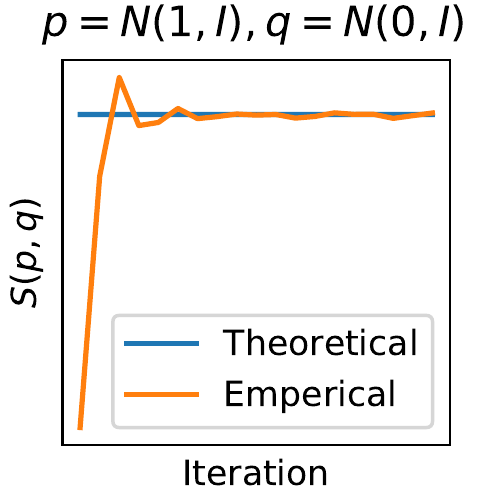}
\includegraphics[width=.2\textwidth]{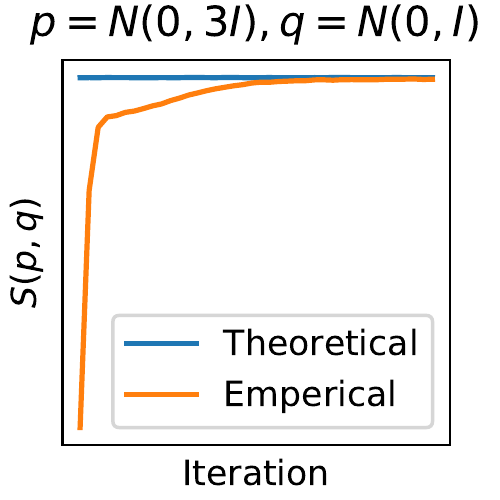}
    \caption{Training a neural net to estimate $\St(p, q)$ on 100-dimensional data. In both cases, a near-optimal critic is learned and the true discrepancy is closely approximated.}
    \label{fig:toy_lsd}
\end{figure}

Thus, the optimal critic depends on $\lambda$ only up to a constant multiplier. While this theory is convenient, it does not tell us if we can learn this function given finite data and compute -- requirements for this to be a useful discrepancy for model evaluation and training. In this work, we demonstrate empirically that we can learn this function from finite data on a wide variety of datasets and models. In Figure \ref{fig:toy_lsd} we show that the theoretically optimal critic can be recovered while estimating the Stein Discrepancy between 100 dimensional Gaussian distributions.


We achieve similar results on more complicated distributions such as RBMs and Normalizing Flows trained on image data. See Sections \ref{sec:rbm_eval} and \ref{sec:flow_eval} for details.

\paragraph{Efficient Estimation}
The $\text{Tr}( \nabla_x f(x) )$ term in Eq.~\ref{eq:critic_obj} is expensive to compute, requiring $O(D)$ vector-Jacobian products (or ``backward-passes''), where $D$ is the dimension of the data.
Fortunately, an efficient, unbiased estimator exists, known as Hutchinson's estimator~\citep{hutchinson1990stochastic}. This estimator has been widely used in the machine learning community in recent years~\citep{grathwohl2018ffjord, tsitsulin2019shape, han2017approximating}. The estimator, which requires only one vector-Jacobian product to compute, is the single-sample Monte-Carlo estimator derived from the following identity:
\begin{align}
    \text{Tr}(\nabla_x f(x)) = \E_{N(\epsilon|0, 1)}\left[\epsilon^T \nabla_x f(x) \epsilon \right]
\end{align}
which can be computed efficiently since $\epsilon^T \nabla_x f(x)$ is a vector-Jacobian product. Thus, during the critic-learning phase, we replace the $\LSD$ objective (Equation \ref{eq:critic_obj}) with
\begin{align}
    \rm{\bf{LSDE}}(f_\phi, p, q) = \E_{p(x)N(\epsilon|0, I)}[&\nabla_x \log q(x)^T f_\phi(x)\nonumber\\
    &+ \epsilon^T \nabla_x f_\phi(x) \epsilon]
    \label{eq:efficient_critic_obj}
\end{align}
an \textbf{E}fficient $\LSD$ version which can be estimated and optimized on mini-batches of data sampled from $p(x)$ for the same asymptotic time cost as evaluating $f$ and $\nabla_x \log q(x)$ once on each sample in the mini-batch.

\section{Model Evaluation with LSD}

Now that we have a procedure for estimating the Stein Discrepancy, we provide two applications related to model evaluation: comparing the performance of models on held-out evaluation data, and goodness-of-fit testing.

\subsection{Model Comparison}
\label{sec:crit}
In this setting we are given two or more unnormalized models $\{q_i\}_{i=1}^M$, and a finite set of samples $\{x_i\}_{i=1}^n$ from the target distribution $p(x)$ which $\{q_i\}_{i=1}^M$ are trying to approximate. The goal is to estimate which $q_i$ approximates the data samples better. 

We phrase this evaluation task as a standard learning problem.
We split the data into training, validation, and testing sets, and train a critic $f_{\phi_i}$ on model $q_i$, using the validation data to do model selection.
We then compute the $\LSD$ on the test data for each model and score the models by this value. 

Due to the nature of our training objective, the variance of the $\LSD$ must be taken into consideration when monitoring for over-fitting. We propose an uncertainty-aware model selection procedure where we choose the model which maximizes $\mu - \sigma$ where $\mu$ and $\sigma$ are the sample mean and standard deviation of the $\LSD$ on the validation data. We provide a more in-depth discussion in Appendix \ref{ap:validation}.

Our approach is summarized in Algorithm~\ref{alg:lsd_comp}.
See Sections~\ref{sec:rbm_eval},~\ref{sec:flow_eval} for experimental result for experimental results, where
we find that $\LSD$ can scale to complicated models
of high-dimensional data such as images.

\renewcommand{\algorithmicrequire}{\textbf{Input:}}
\renewcommand{\algorithmicensure}{\textbf{Output:}}

\begin{algorithm}
\caption{LSD Model Comparison}\label{alg:lsd_comp}
\begin{algorithmic}[h]
\REQUIRE Critic architecture $f_\phi$, models $\{q_i\}_{i=1}^M$, data $\xx = \{x_i\}_{i=1}^n$, $\LT$ regularization hyperparameter $\lambda$
\ENSURE Estimated Stein Discrepancies, $\{\St(p, q_i)\}_{i=1}^M$

\STATE Split $\xx$ into $\xx_{\text{train}}$, $\xx_{\text{val}}$, and $\xx_{\text{test}}$
\FOR{model $q_i$ in $\{q_i\}_{i=1}^M$}
   \STATE find $\phi_i = \text{argmax}_{\phi} \text{LSDE}(f_\phi, \xx_{\text{train}}, q_i) - \mathcal{R}_\lambda(f_\phi)$ (using $\xx_{\text{val}}$ for model selection)
   
   $s_i = \text{LSD}(f_{\phi_i}, \xx_{\text{test}}, q_i)$
\ENDFOR
\STATE Rank models in increasing order of $s_i$.
\end{algorithmic}
\end{algorithm}

\newcommand{\s}[2]{\nabla_x \log #1(x)^T #2(x) + \text{Tr}(\nabla_x #2(x))}

\subsection{Goodness-Of-Fit Testing}
\label{sec:ht}
Given an unnormalized model $q(x)$ and a finite set of samples $\{x_i\}_{i=1}^n$ from an unknown distribution $p(x)$, the problem of Goodness-Of-Fit (GoF) testing asks us to decide between two hypotheses:
\vspace{-.1cm}
\begin{align}
    H_0: p = q, \qquad H_1: p \neq q.\nonumber
\end{align}
GoF is a standard problem in statistics. An ideal hypothesis test will reject $H_0$ whenever $p \neq q$, even if $p$ and $q$ are quite similar. In settings relevant to the machine learning community, GoF becomes particularly challenging as we often deal with very complicated, high-dimensional data.


We use $\LSD$ to develop a procedure for solving GoF problems.
Assuming we are given a critic $f$, we define $s^q_f(x) = \s{q}{f}$. Following Equation \ref{eq:stein_id}, this problem reverts to a test of
\begin{align}
    H_0: \E_{p(x)}\left[s^q_f(x) \right] = 0 \qquad H_1: \E_{p(x)}\left[s^q_f(x)\right] \neq 0\nonumber
\end{align}
which is a simple one-sample location test. This test is carried out by computing the statistic $t = \sqrt{n}\frac{\mu_s}{\sigma_s}$ where $\mu_s$ and $\sigma_s$ are the sample mean and standard deviation of $s_f(x)$ evaluated over the dataset.
For sufficiently large $n$, we have $t \sim N(0, 1)$ under $H_0$, thus for a given test confidence $\alpha$ we should reject $H_0$ if $t < \Phi(1 - \alpha)$ where $\Phi$ is the inverse CDF of the standard Normal distribution. 

As above, to obtain the critic, we parameterize it as a neural network $f_\phi$ and train its parameters on a subset of the given data. In Section~\ref{sec:crit} we were interested in estimating $\St(p, q)$ directly. Here we are only interested in correctly choosing $H_0$ or $H_1$, so we should optimize $f_\phi$ to minimize the probability we accept $H_0$ given $H_1$ is true. The inverse of this quantity is known as the \emph{test power}.
Given that the test statistic $t$ is asymptotically normal under both $H_1$ and $H_0$ we can follow the same argument as \citet{sutherland2016generative, jitkrittum2017linear}, and maximize the test power by optimizing
\begin{align}
    \mathcal{P}(f_\phi, p, q) = \frac{\E_{p(x)}[s^q_{f_\phi}(x)]}{\sigma_{p(x)}[s^q_{f_\phi}(x)]}.
    \label{eq:ht_obj}
\end{align}

We split our finite sample set into disjoint sets $\x_\text{train}, \x_\text{val}$, and $\x_\text{test}$, using $\x_\text{train}$ for training and $\x_\text{val}$ for model selection.
Then, given our learned $f_\phi$ we run the test on $x_\text{test}$. A summary of our test can be seen in Algorithm \ref{alg:ht}. As before, we use Hutchinson's estimator for the trace term in $s^q_f(x)$. Because this increases the variance of $s^q_f(x)$, optimizing the objective maximizes a lower-bound on the test power.

Experimental details and results can be found in Section~\ref{sec:gof}.

\begin{algorithm}
\caption{LSD Goodness of Fit Test}\label{alg:ht}
\begin{algorithmic}[h]
\REQUIRE Critic architecture $f_\phi$, model $q$, $\xx = \{x_i\}_{i=1}^n$, $\LT$ regularization hyperparameter $\lambda$, confidence $\alpha$
\ENSURE Decision whether $p = q$
\STATE Split $\xx$ into $\xx_{\text{train}}$, $\xx_{\text{val}}$, and $\xx_{\text{test}}$
\STATE maximize $\phi = \text{argmax}_{\phi} \mathcal{P}(f_\phi, \xx_{\text{train}}, q)$ (using $\xx_{\text{val}}$ for model selection)
\STATE Compute $t = \sqrt{n} \frac{\E[s(\x_{\text{test}})]}{\sigma[s(\x_{\text{test}})]}$
\STATE If $t > \Phi(1 - \alpha)$ reject $H_0$, else accept $H_0$
\end{algorithmic}
\end{algorithm}

\section{Training Unnormalized Models with LSD}
\label{sec:lsd_train}
Above we have developed a method which can effectively quantify the quality of fit of an unnormalized model to a fixed set of datapoints. We now extend this method to develop a way of learning the parameters of an unnormalized model to best fit the data.
Unlike sampling based approaches, we produce an objective that is 0 in expectation if and only if the data and model distributions are equivalent, given our critic is suitably expressive. Ideally, we will minimize
\vspace{-.1cm}
\begin{align}
    \mathcal{L}(\theta) = \sup_\phi \mathbf{LSD}(f_\phi, p,  q_\theta) - \mathcal{R}_\lambda(f_\phi)
    \label{eq:ideal_critic_obj}
\end{align}
with respect to $\theta$ but this is infeasible due to the supremum on the right-hand-side.
Instead, we follow recent work on minimax optimization~\citep{goodfellow2014generative} and iteratively update the model $q_\theta(x) = \exp(-E_\theta(x))/Z$ and the critic function $f_\phi$ in an alternating fashion where the critic is trained to more accurately estimate the Stein Discrepancy and the model is trained to minimize the critic's estimate of the discrepancy. The proposed training algorithm is summarized in Algorithm \ref{alg:lsd_train}.

\begin{figure}[t]  
\centering
\includegraphics[width=.47\textwidth, trim=10mm 15mm 10mm 3.6mm]{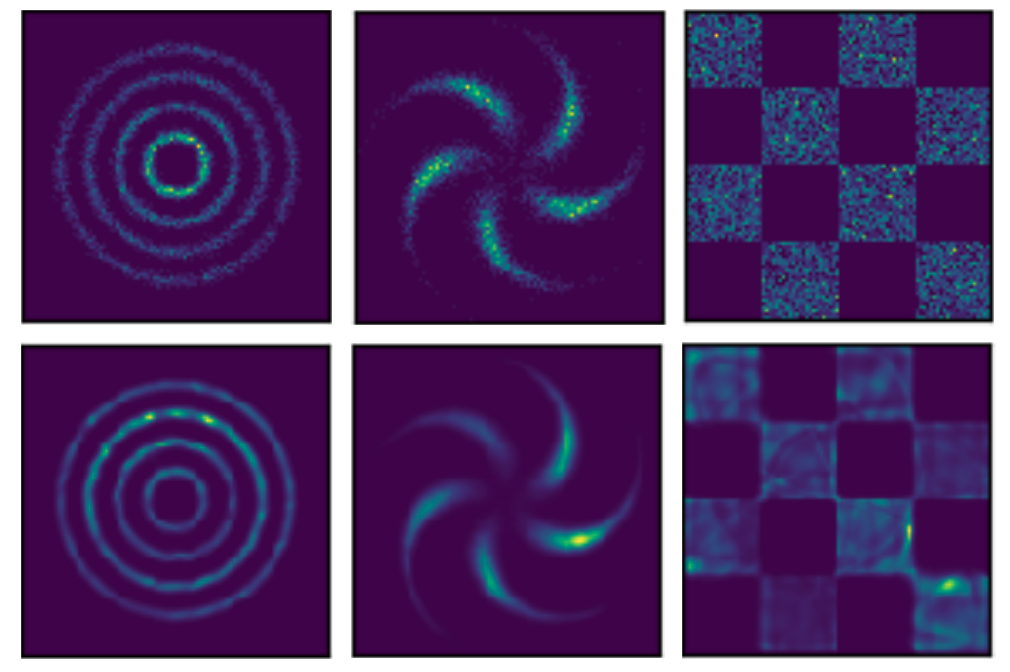}
    \caption{Density models trained using LSD. Top: Data. Bottom: Learned densities. }
    \label{fig:toy_2}
\end{figure}

Our approach can effectively train unnormalized models on complicated high-dimensional datasets 
without requiring samples of the model -- only needing access to the model's score function. In contrast, score matching (which also does not require sampling), requires the Hessian of $\log q_\theta$ for training. For many distributions of interest this computation can be numerically unstable leading to training issues (see Section~\ref{sec:ica} for more details). Further, optimizing functions implicitly through their higher-order derivatives can be challenging as they are often sparse or discontinuous. See Figure \ref{fig:toy_2} for examples of toy densities trained with $\LSD$.

\begin{algorithm}
\caption{LSD Training}\label{alg:lsd_train}
\begin{algorithmic}[h]
\REQUIRE Critic architecture $f_\phi$, model $q_\theta$, data $\xx = \{x_i\}_{i=1}^n \sim p(x)$, $\LT$ regularization hyperparameter $\lambda$, training iterations $T$, critic training iterations $C$ 
\ENSURE Parameters $\theta$ such that $q_\theta \approx p$
\FOR{$T$ iterations}
   \FOR{$C$ iterations }
   \STATE Sample mini-batch $\xx'$
   \STATE Update $\phi$ with $\nabla_\phi \left(\text{LSDE}(f_\phi, \xx', q_\theta) - \mathcal{R}_\lambda(f_\phi)\right)$
   \ENDFOR
   \STATE Sample mini-batch $\xx'$
   \STATE Update $\theta$ with $-\nabla_\theta \text{LSD}(f_\phi, \xx',  q_\theta)$
\ENDFOR
\STATE Return resulting model $q_\theta$
\end{algorithmic}
\end{algorithm}

\begin{figure*}[h]       
    \includegraphics[width=\textwidth]{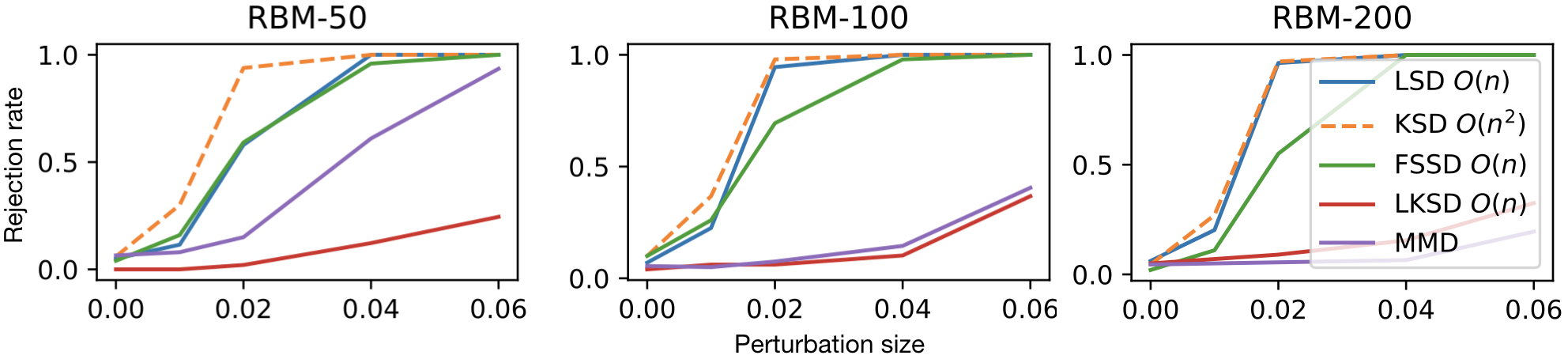}
    \vspace{-.8cm}\caption{Hypothesis testing results. Test confidence $0.05$. Perturbed RBMs of increasing data dimension. Perturbation magnitude on the $x$-axis, rejection rate on the $y$-axis. Number of datapoints $n=1000$. Ideal behavior is a 5\% rejection when perturbation is 0 and close to 100\% rejection otherwise. In high dimensions our linear-time $\LSD$ matches the performance of the quadratic-time KSD.}
    \label{materialflowChart}
    \label{fig:ht}
\end{figure*}

\section{Model Evaluation Experiments}
We run a number of experiments to demonstrate the utility of the $\LSD$. First, we demonstrate that the $\LSD$ can be used to build a competitive test for goodness-of-fit which scales more favorably to high-dimensional data than prior linear-time methods. Next we show that $\LSD$ can be used to provide a fine-grained model evaluation tool which is sensitive to small differences in high-dimensional models.

Unless otherwise stated, all critics $f_\phi$ are MLPs with 2 hidden layers and 300 units per layer.
We use the Swish~\citep{ramachandran2017searching} nonlinearity throughout.

\subsection{Hypothesis Testing}
\label{sec:gof}
We compare our linear-time hypothesis testing method from Section~\ref{sec:ht} with a number of kernel-stein approaches; the quadratic-time Kernelized Stein Discrepancy (KSD), its linear-time variant (LKSD)~\citep{liu2016kernelized}, and the linear-time Finite Set Stein Discrepancy (FSSD)~\citep{jitkrittum2017linear}. We also compare against a kernel-MMD test~\citep{gretton2012kernel} which requires MCMC sampling from the model.
We test these approaches in their ability to determine whether or not a set of samples was drawn from a given Gaussian-Bernoulli Restricted Boltzmann Machine~\citep{cho2013gaussian}. This is an unnormalized latent-variable model
\vspace{-.1cm}
\begin{align}
    p(x, h) = \frac{1}{Z}\exp\left(\frac{1}{2} x^T B h + b^T x + c^T h - \frac{1}{2} ||x||^2 \right)\nonumber\\
    \nabla_x \log p(x) = b - x + B \cdot \texttt{tanh} \left(B^T x + c\right)\nonumber
    \label{eq:rbm}
\end{align}
with parametrs $B, b, c$, whose gradients can be efficiently computed making it an ideal candidate for evaluating methods such as ours. 
We randomly sample the parameters of the model, then draw a set of $n = 1000$ samples. We then perturb the weights of the model with Gaussian noise of standard deviation in $[0, 0.01, 0.02, 0.04, 0.06]$ and our tests must determine if the samples were drawn from this model. We perform this test with RBMs of increasing visible dimension $x$ and hidden dimension $h$ from $x, h = \{(50, 40), (100, 80), (200, 100)\}$.

As can be seen in Figure~\ref{fig:ht}, our proposed hypothesis test performs comparably to the linear-time kernel methods at 50 dimensions and begins to dominate those approaches as dimensionality is increased, matching the performance of the quadratic-time test (we note that this quadratic-time test takes much longer to run than our linear-time method). We also see that while the MMD test performs comparably at 50 dimensions, as we increase dimensionality the performance quickly drops below all of the Stein approaches. This result further supports our claim that MCMC sampling should not be relied upon in high dimensional settings. 

We run further experiments to empirically verify the normality of our test statistic under $H_0$. These results can be found in Appendix \ref{app:test_stat}.

\subsection{RBM Evaluation}
\label{sec:rbm_eval}
We now demonstrate $\LSD$'s ability to rank and evaluate the fit of unnormalized models on fixed test data. Again, we experiment with Gaussian-Bernoulli RBMs. As above we randomly initialize an RBM, draw $n = 1000$ samples, perturb its weights with increasing Gaussian perturbations, and then
and provide a score for each model. This score should increase as the perturbation becomes larger, starting at 0 when $p = q$. As above, we experiment with RBMs of dimension 50, 100, and 200. 

We compare $\LSD$ (which approximates the Stein Discrepancy over $\mathcal{L}_2$ functions) with a linear-time and a quadratic-time estimate of the KSD using RBF Kernels with learned bandwidth. We also compare with the theoretical upper-bound on the $\LSD$ from Equation \ref{eq:bound}. 

\begin{figure*}[h!]       
    \includegraphics[width=\textwidth]{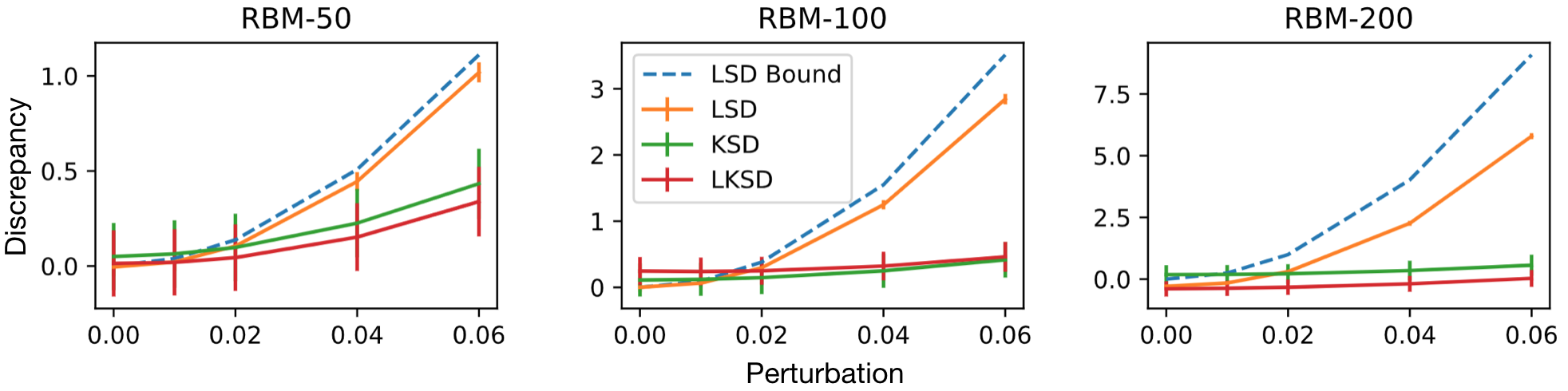}
    \vspace{-.8cm}\caption{Model evaluation results for Gaussian-Bernoulli RBMs. Predicted discrepancy should increase as perturbation increases. $\LSD$ reliably increases, approaching the theoretical ground-truth value. $\LSD$ provides much more certain results than both linear-time and quadratic-time kernel-based approaches.}
    \label{materialflowChart}
    \label{fig:crit}
\end{figure*}

We test all approaches using $n = 1000$ samples as this was the largest $n$ that could feasibly be used for the quadratic-time KSD.
In all dimensions, we find that $\LSD$ is a much stronger measure of model quality.
It is able to closely match the theoretical upper bound even in higher dimensions.
Further, it is much more discriminative for small-scale perturbations than the kernel methods. Results can be seen in Figure~\ref{fig:crit}. We plot the reported scores with error-bars indicating standard deviation of the sample mean $\frac{\sigma}{\sqrt{n_\text{test}}}$.


\subsection{Normalizing Flow Evaluation}
\label{sec:flow_eval}

For the $\LSD$ to be an effective discrepancy for use on problems of scale, it must be able to distinguish between relatively similar, high-dimensional models.
To assess $\LSD$'s ability to do so, we trained a number of normalizing flow models based on the Glow architecture~\citep{kingma2018glow}. We save these models throughout training and record their log-likelihoods on the MNIST test dataset. We compare the standard likelihood evaluation with the $\LSD$ evaluation procedure. We split the MNIST test set into partitions to  train, validate, and test the $\LSD$ critic. 
This ensures that the scores from $\LSD$ were given access to the exact same data as the likelihood evaluation procedure. We also compare with the objective of Score Matching~\citep{hyvarinen2005estimation} to provide a baseline method which does not require the model's normalizing constant.   
\begin{figure}[H]       
    \centering
    \includegraphics[width=.42\textwidth]{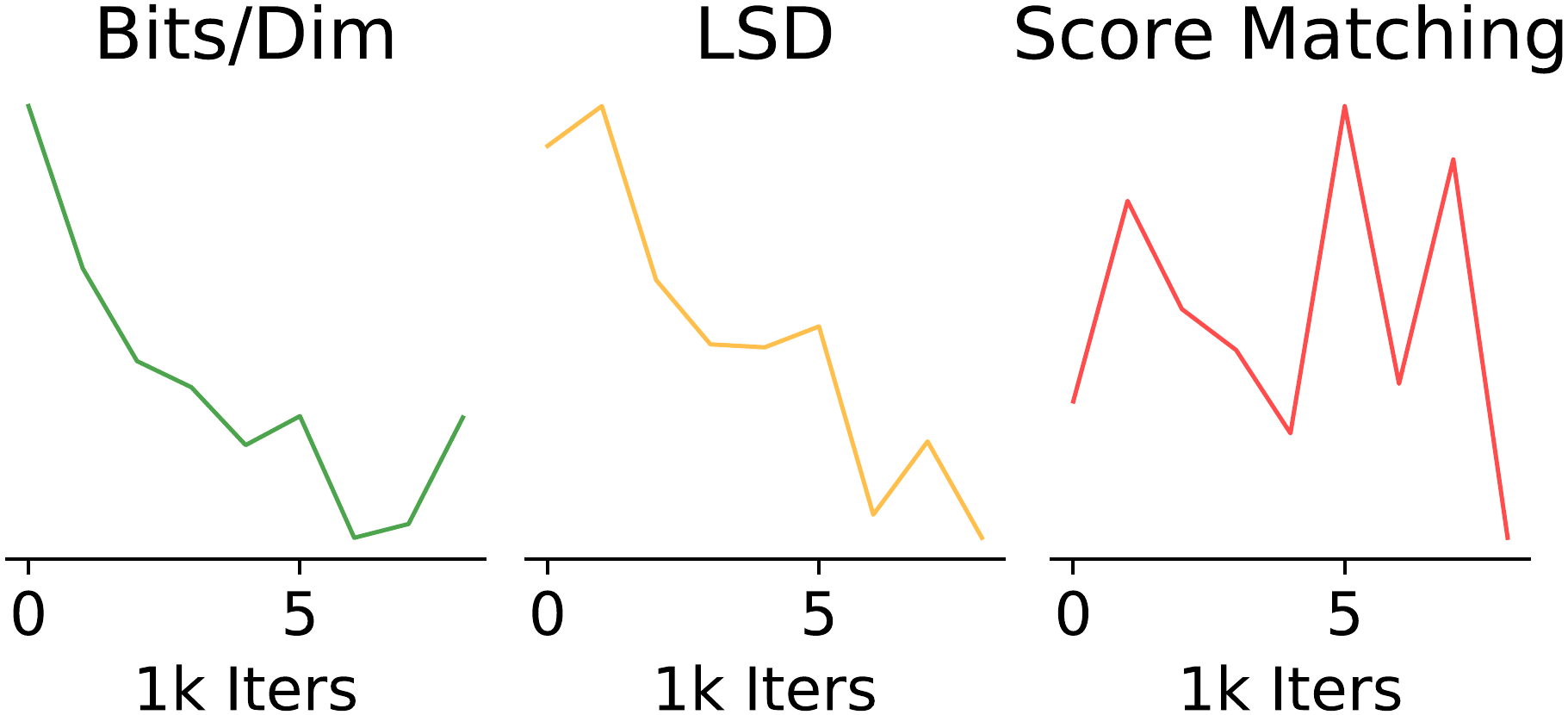}
    \caption{Flow model evaluation. Left to right: Bits/Dim, LSD, Score Matching. Y-axis is removed because scales are not comparable.}
    \label{fig:flow_crit}
\end{figure}
Results can be seen in Figure~\ref{fig:flow_crit}. We find that the score reported by $\LSD$ goes down throughout training, tracking the negative log-likelihood. Conversely, score matching reports a score which is not highly correlated with likelihood, demonstrating that $\LSD$ can be a more effective metric to compare unnormalized models than previous approaches. 


\section{Model Training Experiments}
Here we demonstrate that minimizing $\LSD$ is an effective method for training unnormalized models which scales to high dimensional data more effectively than previously proposed approaches. We have trained deep EBMs on some simple toy 2D densities using $\LSD$ which can be seen in Figure~\ref{fig:toy_2}. Below we present quantitative results on some more challenging problems. 

\begin{figure*}[h]
%
\begin{minipage}{.64\linewidth}
\begin{tabularx}{\textwidth}{c |c c c c c } 
 Method                                &    &  \multicolumn{3}{c}{Data Dimension}    &   \\
                                       & 10 & 20 & 30 & 40 & 50 \\
 \midrule
 Max. Likelihood                       & -10.98 & -18.48 & -21.49 & -23.43 & -25.53 \\ 
 \midrule
 $\LSD$ (Ours)                            & $-10.95$ & $\bold{-18.37}$ & $\bold{-21.23}$ & $\bold{-25.14}$ & $\bold{-25.36}$ \\
 Score Matching                        & -11.13 & -27.20 & -21.48 & NaN    & NaN    \\
 NCE                                   & $\bold{-10.92}$ & -22.52 & -30.33 & -55.53    & -73.62 \\
 CNCE                                  & -11.00 & -18.77  & -24.47 & -37.64 & -36.31 \\
 \bottomrule
\end{tabularx}
\end{minipage}
\hspace{0.1cm}
\vspace{-.18cm}
\begin{minipage}{.33\linewidth}
    \includegraphics[width=1.\textwidth]{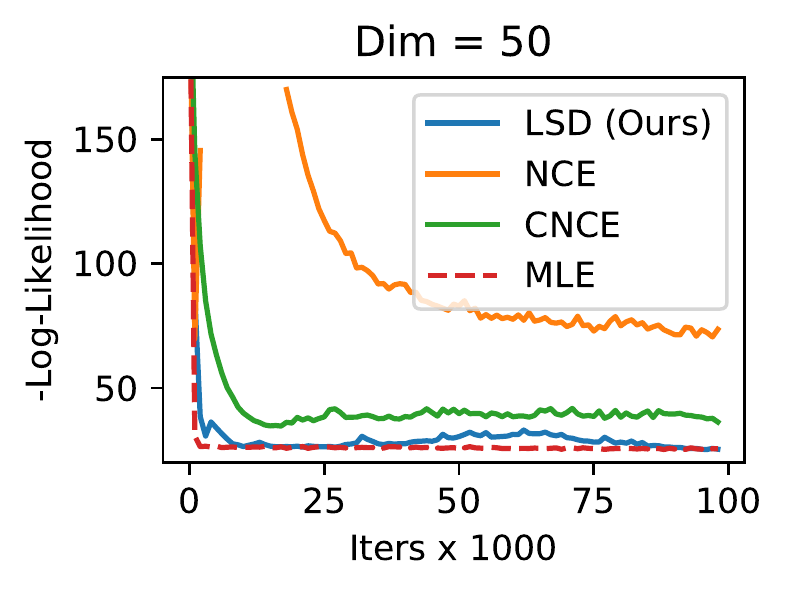}
\end{minipage}
\caption{Linear ICA training results. Left: Table comparing test set log-likelihoods of linear ICA models with various training methods. $\LSD$ closely tracks the performance maximum likelihood training while other unnormalized methods fall behind or diverge. Right: Learning curves for 50-dimensional ICA. While slower to converge than maximum likelihood, $\LSD$ cleanly converges to the same result.} 
\label{fig:ica}
\end{figure*}

\subsection{Linear ICA}
\label{sec:ica}
The linear Independent Components Analysis (ICA) model is commonly used to quantitatively evaluate the performance of methods for training unnormalized models~\citep{gutmann2010noise, hyvarinen2005estimation, ceylan2018conditional}. It consists of a simple generative process
\begin{align*}
    z \sim \text{Laplace}(0, 1), \qquad x = Wz
\end{align*}
where the model parameter $W$ is a $D \times D$, non-singular mixing matrix. The log-density of a datapoint $x$ under this model is
\begin{align}
    \log p(x; W) = \log p_z\left(W^{-1} x\right) - \log |W|
\end{align}
where $p_z(\cdot)$ is the PDF of the Laplace(0, 1) distribution.



We train linear ICA models on randomly sampled mixing matrices using various methods and compare their performance as dimension $D$ is increased from 5 to 50. To provide an upper-bound on possible performance, we train with brute-force Maximum Likelihood (ML) (which requires inverting the parameter matrix at each iteration). We compare $\LSD$ with other approaches to train unnormalized models; Noise-Contrastive Estimation (NCE)~\citep{gutmann2010noise}, Conditional Noise-Constrastive Estimation (CNCE)~\citep{ceylan2018conditional}, Score Matching (SM)~\citep{hyvarinen2005estimation}.

Results can be seen in Figure \ref{fig:ica}. We find that for smaller $D$ these methods behave comparably, and all arrive at a similar solution to maximum likelihood. For $D$ above 20, other unnormalized training methods fail to achieve the same level of performance as ML while $\LSD$ does. 
For $D$ above 30, SM quickly diverged due to instabilities that arose with computing the second derivatives of the Laplace log-density. Our approach only requires access to the first derivatives of the model and the critic, avoiding these issues. Further experimental details can be found in Appendix \ref{sec:ica_details}.


\subsection{Preliminary Image Modeling Results}
We now show that $\LSD$ can be used to train much more expressive unnormalized models. We train deep energy-based models with $\LSD$ on MNIST and FashionMNIST. The models take the form of $q_\theta(x) = \exp(-E_\theta(x))/Z$,
where the energy-function
$E_\theta(x)$ is parameterized by a neural network with a single output variable. We parameterize the critic $f_\phi$ with a neural network which maps $R^D \rightarrow R^D$.

\begin{figure}[t]    
    \centering
    \includegraphics[width=.22\textwidth]{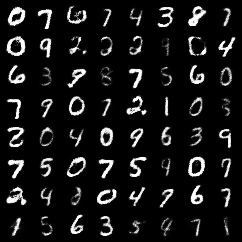}
    \includegraphics[width=.22\textwidth]{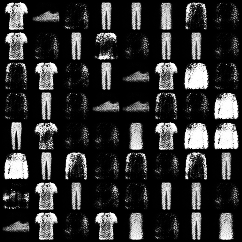}
    \caption{Samples from deep EBMs trained with LSD. \emph{Left}: MNIST, \emph{Right}: FashionMNIST.}
    \label{fig:mnist_samples}
    \vspace{-.5cm}
\end{figure}

After training, we draw approximate samples with a tempered SGLD sampler. These can be seen in Figure \ref{fig:mnist_samples}. Without the use of a sampler, $\LSD$ has trained a model which captures the various modes of the data distribution and can be used to draw compelling samples.
Of course, these MCMC samples are susceptible to all of the issues we have mentioned previously in this work. Their quality is tightly coupled with the post-hoc sampler's parameters and we make no claims that these represent true samples from our models. 


While these samples are not state-of-the-art, they do showcase the ability of $\LSD$ to scale to high-dimensional problems. Scaling further to large natural images will potentially require tricks from the GAN literature, the use of convolutional architectures, and more advanced samplers. We leave this for future work. We refer the reader to Appendix~\ref{ap:im_details} for experimental details, and Appendix~\ref{sec:im_extra} for further samples and experiments using $\LSD$ to train RBMs on image datasets. 



\section{Related Work}
\paragraph{Sampler-Free EBM Training}
Ours is not the first method to train unnormalized densities without MCMC sampling. Score matching (SM)~\citep{hyvarinen2005estimation} matches the gradients of the model density to the data density, circumventing the computation of the normalizing constant. Unlike our approach, SM requires the computation of the unnormalized density's Hessisan trace, which can be expensive and unstable (see our ICA results in Section~\ref{sec:ica}). An alternative interpretation of SM also exists known as Denoising Score Matching~\citep{vincent2011connection} which approximates the SM objective without the Hessian term. This has been applied to deep EBMs~\citep{saremi2018deep} and recently scaled to high resolution image data~\citep{song2019generative, li2019annealed} with compelling results.

These methods attempt to minimize the Fisher Divergence $\mathbf{F}(p, q) = E_{p(x)}[||\nabla_x \log p(x) - \nabla_x \log q(x)||^2_2]$, which can be viewed a special case of the Stein Discrepancy with a specific, fixed critic function~\citep{liu2016kernelized}. \citet{barp2019minimum} showed that other EBM training methods (such as approximate maximum likelihood) can be viewed as minimizing a Stein Discrepancy with respect to a different class of critics.  


\paragraph{Approximating Stein Discrepancies}
A second highly relevant prior contribution is the Stein Neural Sampler~\citep{hu2018stein}.
which presents a setup very similar to our own but with the opposite motivation. In their work, one is given an unnormalized density $q(x)$ (such as the posterior of a Bayesian model) and seeks to learn an implicit sampler $x = g_\theta(z)$ such that $x\sim q(x)$. This is achieved through a training procedure like ours presented in Section~\ref{sec:lsd_train} but with $q(x)$ fixed, and the gradients of $\LSD$ back-propagated back through the samples, to the parameters of the sampler. 

Our work benefited significantly from the theoretical results of their work, but their method was not particularly scalable due to the brute-force evaluation of a Jacobian trace. We solve this issue with Hutchinson's estimator (Eq.~\ref{eq:efficient_critic_obj}). We are particularly excited about this line of work and believe there is much promise to further scale these sampling methods in this way. 

\paragraph{Variational Inference}
Computing and minimizing Stein Discrepancies has been studied in the context of variational inference for both sampling~\citep{liu2016stein} and amortized infernece. In particular, \citet{ranganath2016operator} introduce an objective for inference similar to LSD but utilize a different class of critic function and not dot apply many of the techniques we used for scalability. We expect using our class of critics and these techniques could improve the performance of their approach.

\paragraph{GANs and Adversarial Optimization}
Generative adversarial networks (GANs)~\citep{goodfellow2014generative} (and specifically, their extension Wasserstein GANs~\citep{arjovsky2017wasserstein}) have many similarities to LSD.  They both train a model by minimizing a discrepancy between the model distribution and the data distribution which is defined through a critic function. Both approaches parameterize this critic with a neural network and train this critic online with the model in an alternating fashion. The main difference lies in the properties of the model; the WGAN trains an implicit sampler while $\LSD$ trains an unnormalized density model.

GANs have successfully scaled to very high dimensional datasets~\citep{brock2018large} and their success has motivated considerable progress in minimax optimization. Due to the similarity of the approaches, we are excited about using these techniques to scale $\LSD$ to more challenging problems than we have attacked in this work. 

\section{Conclusion}

In this work we have presented $\LSD$, a novel and scalable way to train and evaluate unnormalized density models without the need for sampling or  kernel-selection heuristics. $\LSD$ outperforms state-of-the-art methods for high-dimensional hypothesis testing. We show that $\LSD$ tracks likelihood much better than common objective functions used to train unnormalized density models. Finally, we show that $\LSD$ can be used to train unnormalized density models efficiently in high dimensions.

\section{Acknowledgements}
We thank Renjie Liao and Murat Erdogdu for useful discussion. Resources used in preparing this research were provided, in part, by the
Province of Ontario, the Government of Canada through CIFAR, and companies
sponsoring the Vector Institute (\url{www.vectorinstitute.ai/\#partners}).

\bibliography{example_paper}
\bibliographystyle{icml2020}

\newpage
\appendix

\section{Model Selection}
\label{ap:validation}
Here we elaborate on some considerations that must be taken when training the $\LSD$ on finite data. Like any over-parameterized machine learning model, our critic networks are susceptible to over-fitting. 

Our training objective repeated is:
\begin{align}
    \mathbf{LSD}(f_\phi, q, p) = \E_{p(x)}[\nabla_x \log q(x)^T f_\phi(x) + \text{Tr}(\nabla_x f_\phi(x))].\nonumber
    \label{eq:critic_obj_repeat}
\end{align}

If the regularization parameter $\lambda$ is not set large enough, then this objective can be increased on the finite training data by simply adding a scalar multiplier onto the output of $f_\phi$. This has the effect of increasing the variance of the $\LSD$. Since we would like to select the model which maximizes the expectation of the $\LSD$, a natural validation statistic would simply be the mean of the $\LSD$ on the validation set. Unfortunately, as the variance of the $\LSD$ increases, so does the variance of this statistic making this statistic an increasingly unreliable predictor of the model's true performance. This means it is possible for this validation statistic to continue to increase as the model is over-fitting.

To combat this behavior, we propose a variance-aware validation statistic which is simply $ \mu - \sigma$ or the mean of the $\LSD$ minus its standard deviation computed on the validation data. This statistic will decrease as the estimator variance increases, solving the issue. 

\begin{figure}[h]       
    \includegraphics[width=.5\textwidth]{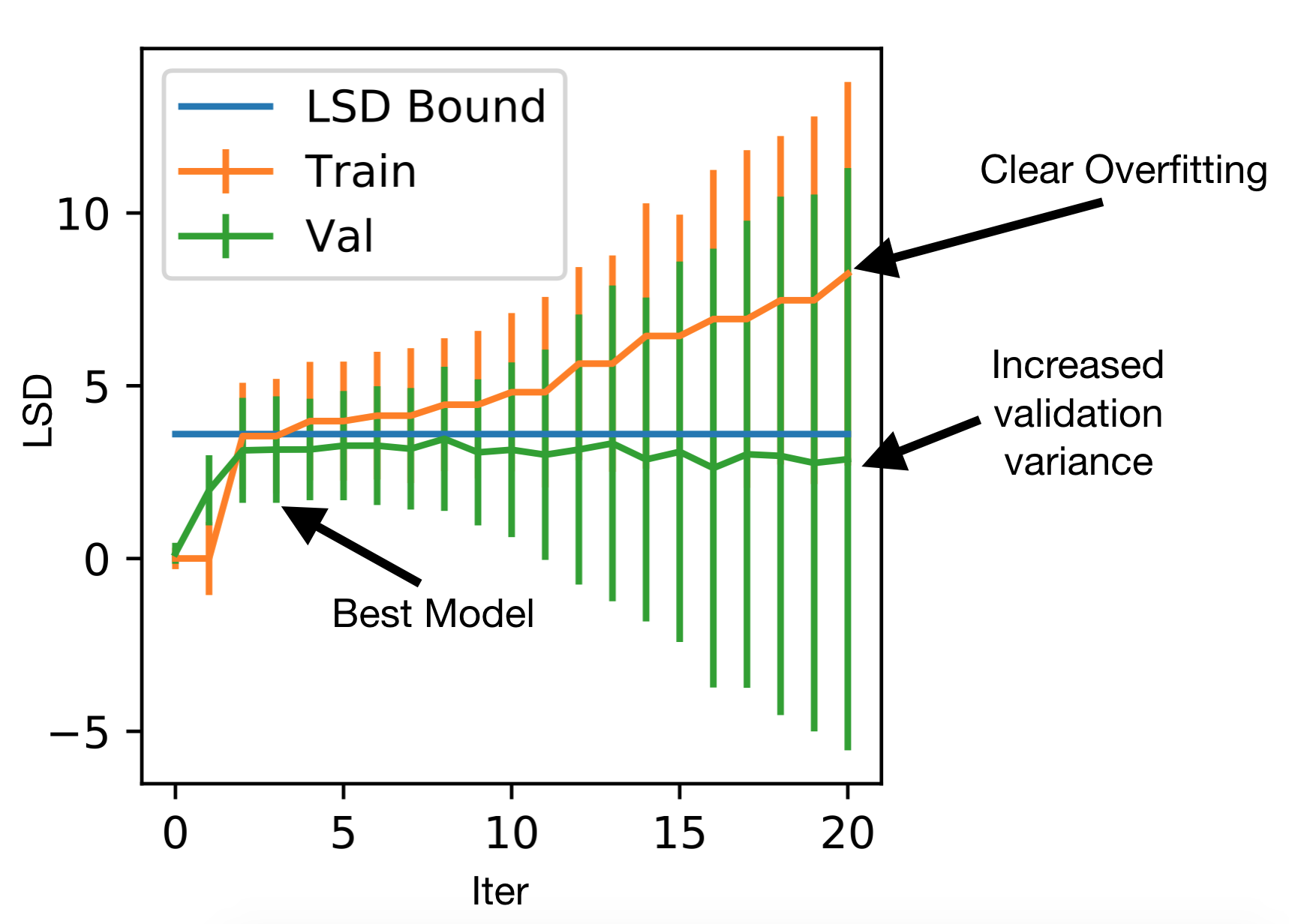}
    \caption{Visual motivation of our variance-aware model selection procedure. Plots of mean and standard deviation of the $\LSD$ computed on training and validation data.}
    \label{fig:varfig}
\end{figure}

We demonstrate this visually in Figure \ref{fig:varfig}. We train a critic to estimate the $\LSD$ between data sampled from a 10-dimensional $N(0, 1)$ and a $N(1, 1)$ model $q$. The training dataset consists of 100 examples, so a flexible neural network critic will certainly over-fit. We see that as the model trains, the $\LSD$ evaluated on the 100-example training set quickly passes the theoretical bound (which is the true Stein Discrepancy), indicating over-fitting. The variance of the $\LSD$ validation data increases with over-fitting. We see that the best model is selected by our variance-aware model selection procedure.

\subsection{Hypothesis Testing}
We note that the above model selection procedure is \emph{not required} when training critics for hypothesis testing. The test power objective (Equation \ref{eq:ht_obj}) explicitly minimizes its  variance so standard model selection may be used. 



\section{Experimental Details}
\subsection{Hypothesis Testing}
We initialize RBMs following \citet{liu2016kernelized, jitkrittum2017linear}. We select the entries of $B$ uniformly from $\{+1, -1\}$ and we sample $b$ and $c$ from a standard Gaussian distribution. To test in a setting where the quadratic time test can be used for comparison, we use $n = 1000$ samples for our test. 

For the adaptive kernel methods, we use 200 of these samples for learning the kernel parameters and use the remaining 800 samples to compute the test statistic. Given that $\LSD$ has many more parameters to fit, we use 800 samples for training, 100 samples for validation and 100 samples for running the final test. 

We parameterize the critic $f_\phi$ as a 2-layer MLP with 300 units per layer and use the Swish nonlinearity. Due to the small amount of training data available, we regularize the model with dropout~\citep{srivastava2014dropout} and weight decay of strength .0005. We train using the Adam optimizer~\citep{kingma2014adam} for 1,000 iterations. We do not use mini-batches, each iteration uses the entire training set. Every 100 iterations, we compute $\mathcal{P}_\phi(\x_\text{val})$ (Equation \ref{eq:ht_obj}) on the validation data and select the critic with the highest value. We then compute the test statistic using this critic. For all tests we let $\lambda = .5$

Rejection rates presented are the average over 200 random tests.



\subsection{RBM Evaluation}
Experimental setup, (train/validation/test) splits, and models are the same as above but the objectives for training and testing differ. The hypothesis testing models ($\LSD$ and baselines) are trained to optimize test power which is the ratio of the mean of the estimated discrepancy over its standard deviation. This is the optimal objective when we wish to make a binary choice. Here we are interested in ranking models. Thus we train the baselines to maximize the discrepancy they report. This means maximizing Equation \ref{eq:ksd} with respect to the kernel parameters. For the $\LSD$ models, we maximize Equation \ref{eq:efficient_critic_obj}. For all models we let $\lambda = .5$.

We must be aware of estimator variance when doing model selection. We follow the procedure outlined in Appendix \ref{ap:validation} to do model selection using our validation set. Over-fitting is not an issue with the kernel methods, so this is not done for the baselines. 

\subsection{Flow Evaluation}
We train a small normalizing flow models based on the Glow architecture~\citep{kingma2018glow}. The flow is an additive multi-scale architecture, uses reverse shuffle dimension splitting, actnorm, has 3 stages with 4 blocks per stage. Each block has 3 convolutional layers with a kernel size of: $(3\times3,1\times1,3\times3)$ respectively and 128 channels. The flow was trained with the Adam optimizer, a learning rate of $.0001$. We checkpoint the model 10 times throughout training, once every 1000 iteration. The models get between 2.0 and 1.8 bits/dim. 

We take the 10,000 examples in the MNIST test set and split them into subsets for training, validation, and final score estimation. We use 8000, 1000, and 1000 examples for this, respectively. The critic was a 2-layer MLP with 1000 units per layer and used the Swish nonlinearity. We train for 100 passes through the data using a batch size of 128, validating after every pass. 

\subsection{Toy Densities}
The toy densities in Figure \ref{fig:toy_2} and the critic used to train them were parameterized by a 2-layer MLP with 300 units and Swish nonlinearity. The models were trained with the Adam optimizer~\cite{kingma2014adam} for 10,000 iterations with a step size of $.001$. In both the critic and model, we set the Adam momentum parameters as $\beta_1 = .5$, $\beta_2 = .9$. The critic was updated 5 times for every model update. To ensure that the our model density can be normalized, we multiply the density we parameterize by a Gaussian distribution's density:
\begin{align}
    q_\theta(x) = \frac{\exp(-E_\theta(x))N(x; \mu, \sigma^2)}{Z}
\end{align}
and we also learn the parameters $\mu$ and $\sigma^2$ with our model. We find empirically this Gaussian learns to cover the data and the energy function learns to push down the Gaussian's likelihood at areas away from the data.

\subsection{Linear ICA}
ICA weight matrices were generated as random Gaussian matrices such that their condition number was smaller than the dimension of the matrix. Results shown are mean over 5 random seeds.

To make score matching efficient, we also utilize Hutchinson's estimator when computing the Hessian trace, thus the objective we use to train is
\begin{align}
    \label{eq:ssm}
    \mathcal{J}_\theta(x) &= \frac{1}{2}||\nabla_x \log q_\theta(x)||^2 + \epsilon^T \mathcal{H}_x(\log q_\theta(x)) \epsilon^T\nonumber\\
    \epsilon &\sim N(0, I).
\end{align}
This modified score matching objective has been studied previously and is known as Sliced Score Matching~\citep{song2019sliced}.

All baselines were optimized for test performance. For all methods, the learning rate was selected from $[.001, .0001, .00001]$.

For Noise Contrastive Estimation~\citep{gutmann2010noise}, we fit a Guassian to the training data and use this as our noise distribution.

For Conditional Noise Contrastive Estimation, we must choose the standard deviation of our noise distribution. We treat this as a hyper-parameter and search over values in $[.01, .1, 1. 10]$.

$\LSD$ has two hyper-parameters: the number of critic steps for every model step and the $\LT$ regularizer $\lambda$. We search over 1 and 5 for the critic steps and $[.1, 1. 10.]$ for $\lambda$. For all dimensions, we find that the using 1 critic step works the best. 

Models were all trained for 100,000 steps with the Adam~\citep{kingma2014adam} optimizer. In both the critic and model, we set the Adam momentum parameters as $\beta_1 = .5, \beta_2 = .9$.

\label{sec:ica_details}

\subsection{Image Modeling}
\label{ap:im_details}
Our EBM takes the form
\begin{align}
    q_\theta(x) = \frac{\exp(-E_\theta(x))N(x; \mu, \sigma^2)}{Z}
\end{align}

where $E_\theta$ is a 2-layer MLP with 1000 hidden units per layer using the Swish nonlinearity. We multiply the energy-function's output by a Gaussian density (with learned parameters) which guarantees that our model is can be normalized. The critic architecture is identical but with $28 \times 28 = 784$ output dimensions.

We train models on the MNSIT and FashionMNIST datasets for 100 epochs using the Adam optimizer with a step-size of .0001. In both the critic and model, we set the Adam momentum parameters as $\beta_1 = .5, \beta_2 = .9$. We update the critic 5 times for every model update. We set $\lambda = 10.$

\subsubsection{Sampling}
Model samples are drawn with Stochastic Gradient Langevin Dynamics~\citep{welling2011bayesian}. This is an MCMC sampler with the following update rule:
\begin{align}
    x_0 \sim p(x_0), \qquad  \alpha_t \sim N(0, \epsilon I)\nonumber\\
    x_{t+1} = x_{t} - \frac{\epsilon}{2}\nabla_x E_\theta(x_t) + \alpha_t
\end{align}

which is quite similar to gradient descent with random Gaussian noise added. To use the sampler in this form, a number of choices must be made. We must select an initialization distribution $p(x_0)$, a step-size $\epsilon$ and a number of steps $T$ to run the sampler. Each of these can dramatically affect the distribution of the final sample $x_T$. 

In practice, this sampler can be quite slow to draw samples, so a temperature parameter is added to scale up the impact of the gradient $\nabla_x E_\theta(x_t)$ relative to noise $\alpha_t$. This has the effect of decoupling the step-size from the scale of the Gaussian noise. This changes the update rule to:
\begin{align}
    x_0 \sim p(x_0), \qquad \alpha_t \sim N(0, \sigma^2 I)\nonumber\\
    x_{t+1} = x_{t} - \epsilon\nabla_x E_\theta(x_t) + \alpha_t.
\end{align}
This temperature-scaling approach has been used in most recent work on large-scale EBMs~\citep{nijkamp2019anatomy, grathwohl2019your, nijkamp2019learning, du2019implicit}. With this modified version of the sampler we must select: the initialization distribution $p(x_0)$, the step-size $\epsilon$, the noise scale $\sigma$ and the number of steps $T$ -- further complicating the sampling process.

The samples seen in Figure \ref{fig:mnist_samples} were generated using with $\epsilon = 1.$ and $\sigma = .01$ and the sampler was run for 1000 steps. The samples were initialized to a uniform distribution over the data space.

\subsubsection{Pre-Processing}
The MNIST and Fashion MNIST datasets consist of 28 by 28 images. Each pixel is represented with an integer value in $[0, 1, \ldots, 254, 255]$. We dequantize by adding uniform$(0, 1)$ noise to these pixel values, and scale to the range $(0, 1)$. We then apply a logit transformation ($\log x - \log(1 - x)$) which maps $(0, 1) \rightarrow R$. All training and sampling is done in logit-space and samples are mapped back to their original space with a Sigmoid function for visualization.

\section{Analysis of the LSD Goodness-of-fit Test Statistic}
\label{app:test_stat}
Our analysis of the test statistic used in the LSD GOF test rely on the central limit theorem taking effect giving our statistic a $N(0, 1)$ distribution under the null hypothesis. With $n \ll \infty$ one may wonder if our test statistic still has this property. To explore this we run a number of tests to determine normality of data.

To generate statistic samples for testing we run 200 tests under $H_0$ for RBMs with dimehsion 50, 100, and 200. For each we run a Kolmogorov–Smirnov test and compute a qq-plot. The $p$-values of the Kolmogorov–Smirnov tests are 0.29, 0.68, and 0.45, respectively indicating that for each model size there is little evidence to indicate the statistic is not distributed as $N(0, 1)$. 

\begin{figure*}[t]       
    \centering
    \includegraphics[width=\textwidth]{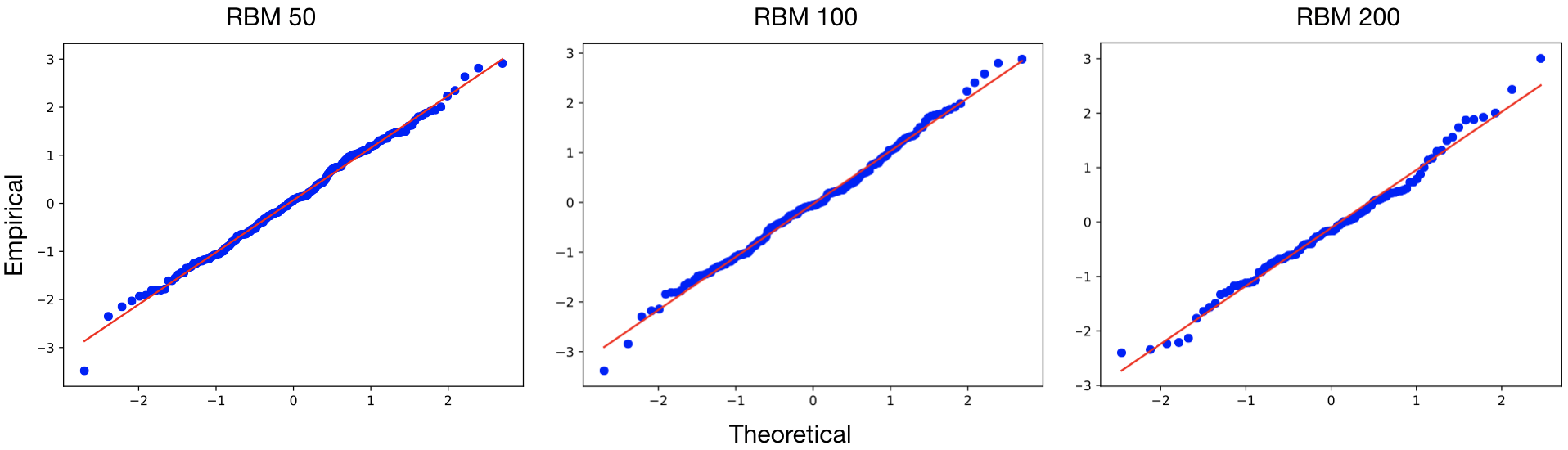}
    \caption{Quantile-Quantile plots of the test statistic compared to $N(0,1)$ for RBMs of dimension 50, 100, 200, left to right.
    When the two distributions are identical, the quantiles will fall along the line $y=x$ (red) given enough samples.  These results indicated that our test's p-values are well-calibrated under the null hypothesis.}
    \label{fig:qq}
\end{figure*}

Next we show the qq-plot which plots the the quantiles of an empirical histogram against the quantiles of the target distribution ($N(0, 1)$ in this case). Ideally the values should lie on the line $y=x$ indicating the empirical CDF aligns with the theoretical CDF. These plots can be found in Figure \ref{fig:qq}. As we see, the empirical distribution of the statistic very closely matches the desired distribution for each model size indicating that our test statistic behaves as the theory would suggest.

\section{Further Image Modeling Results}
\subsection{Additional MNIST Samples}
\begin{figure}[h]       
\centering
    \includegraphics[width=.22\textwidth]{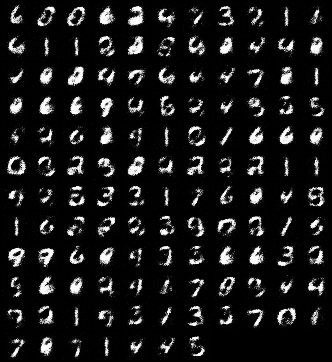}
    \includegraphics[width=.22\textwidth]{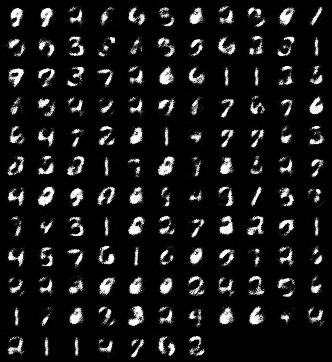}
    \includegraphics[width=.22\textwidth]{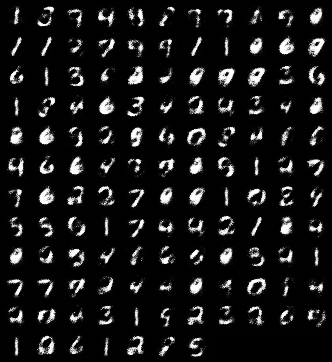}
    \caption{Additional MNIST samples}
    \label{fig:mnist_additional}
\end{figure}

We show additional samples from our MNIST model in Figure~\ref{fig:mnist_additional}. 

\label{sec:im_extra}
\subsection{Exploring Sampler Parameters for Deep EBMs}
\begin{figure}[h]      
\centering
    \includegraphics[width=.22\textwidth]{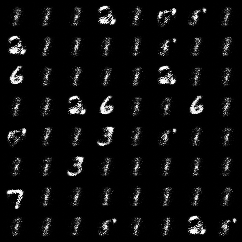}
    \includegraphics[width=.22\textwidth]{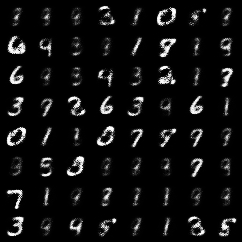}
    \caption{Samples from a deep EBM with different sampler parameters, but the 
    same initialization. Left: $\sigma = .01$, Right: $\sigma = .1$}
    \label{fig:sampler_init}
\end{figure}

Here we demonstrate the impact of the approximate MCMC sampler's parameters on sample quality. In Figure~\ref{fig:sampler_init} we initialize 2 sets of samples to the same value and run SGLD samplers with different noise-scale  parameters 300 steps. Clearly, the samples on the right are more diverse and visually appealing, but this does not tell us that they are \emph{better} samples. 


\begin{figure}[h] 
\centering
    \includegraphics[width=.22\textwidth]{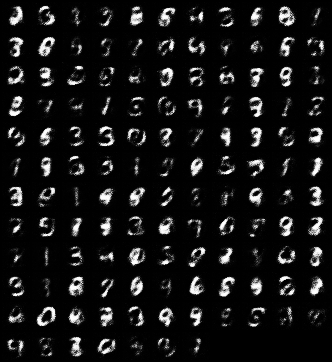}
    \includegraphics[width=.22\textwidth]{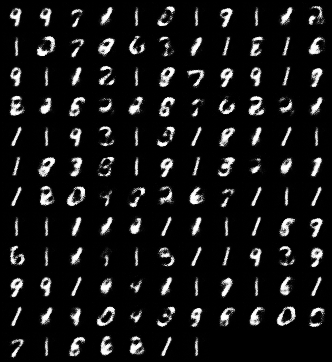}
    \caption{Samples from a deep EBM with different sampler parameters but the same initialization. Left: 30 steps, Right: 1000 steps}
    \label{fig:sampler_steps}
\end{figure}

Next, we demonstrate the impact of the number of sampler steps on sample quality. We initialize two sets of samples to the same value and run a sampler for 30 steps and then 1000 steps. Results can be seen in Figure \ref{fig:sampler_steps}. We can see that initially the samples are quite diverse and as the sampler is run longer, the samples become more clear and high-quality but considerably less diverse. 

We find the choice of sampler parameters has a large impact on the quality and diversity of the resulting samples. Any sample-based evaluation metric would rate each of these sample sets quite differently. Thus these metrics would rate our model differently based on the choice of sampler parameters. Given that these samples are completely separate from our model, this is not ideal behavior.

\subsection{RBMs}
\begin{figure}[h] 
\centering
    \includegraphics[width=.22\textwidth]{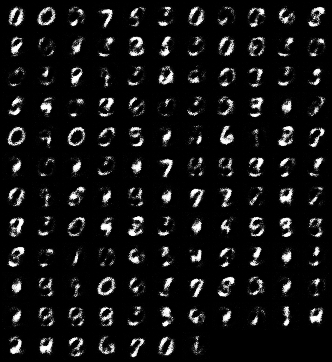}
    \includegraphics[width=.22\textwidth]{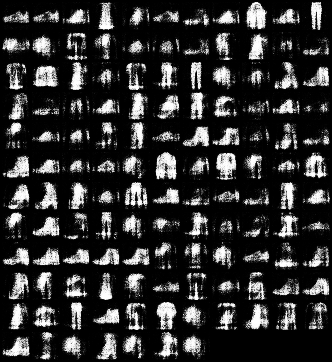}
    \caption{Samples from RBMs trained with LSD. Left: MNIST, Right: FashionMNIST .}
    \label{fig:rbm_samples}
\end{figure}

We find $\LSD$ can be used to train a variety of models -- ICA, deep EBMs and also Gaussian-Bernoulli RBMs. 
Here we train a single-layer RBM with 100 hidden units on MNIST and FashionMNIST. Samples from the models can be seen in Figure~\ref{fig:rbm_samples}. Samples were generated with a Gibbs sampler chain run for 2000 iterations. 

Models are trained for 100 epochs using the Adam optimizer with a step size of .001. We use 5 critic updates per model update and set $\lambda = 10$. We use the same pre-processing as in our deep EBM experiments. 

\end{document}